\documentclass{article}
\usepackage{spconf,amsmath,graphicx,hyperref}
\usepackage{booktabs}

\title{A Training-Free Framework for High-Fidelity Appearance Transfer via Diffusion Transformers}
\name{Shengrong Gu$^{1}$ \quad Ye Wang$^{2}$ \quad Song Wu$^{4 *}$ \quad Rui Ma$^{2}$  \quad Qian Wang$^{4}$ \quad Lanjun Wang$^{3}$   \quad Zili Yi$^{1,5 *\thanks{*Corresponding authors}}$}
  
\address{$1$ School of Intelligence Science and Technology, Nanjing University, Suzhou, China \\
$^{2}$ School of Artificial Intelligence, Jilin University, Changchun, China \\
$^{3}$ School of New Media and Communication, Tianjin University, Tianjin, China \\
$^{4}$ JIUTIAN Research, Bejing, China \\
$^{5}$ State Key Laboratory of Novel Software Technology, Nanjing University, Nanjing, China \\
wusong5533@gmail.com, yi@nju.edu.cn}
\begin{document}
\ninept
\maketitle
\begin{abstract}
Diffusion Transformers (DiTs) excel at generation, but their global self-attention makes controllable, reference-image-based editing a distinct challenge. Unlike U-Nets, naively injecting local appearance into a DiT can disrupt its holistic scene structure. We address this by proposing the first training-free framework specifically designed to tame DiTs for high-fidelity appearance transfer. Our core is a synergistic system that disentangles structure and appearance. We leverage high-fidelity inversion to establish a rich content prior for the source image, capturing its lighting and micro-textures. A novel attention-sharing mechanism then dynamically fuses purified appearance features from a reference, guided by geometric priors. Our unified approach operates at 1024px and outperforms specialized methods on tasks ranging from semantic attribute transfer to fine-grained material application. Extensive experiments confirm our state-of-the-art performance in both structural preservation and appearance fidelity.
\end{abstract}
\section{Introduction}

Recent Diffusion Transformers (DiTs) \cite{peebles2023scalable, esser2024scaling, flux2024} have revolutionized text-to-image generation, producing images with stunning photorealism and detail. A natural next step is to leverage these powerful models for reference-based image editing, where visual attributes like color, texture, or material are transferred from a reference image to a target scene \cite{li2023dreamedit, chen2024anydoor}. This technology holds immense promise for creative applications, from artistic design to virtual prototyping.

\begin{figure}[ht]
  \centering
  % NOTE: Strongly recommend replacing this figure with a new, more impressive one.
  \includegraphics[width=\linewidth]{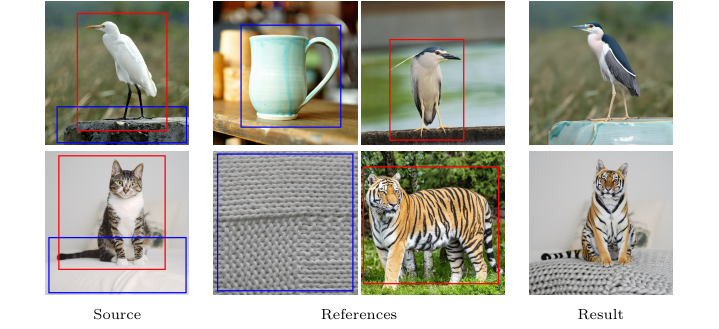} 
  \caption{Our unified, DiT-based framework enables high-fidelity appearance transfer across a wide range of tasks. It can seamlessly transfer semantic attributes (e.g., a tiger's pattern to a cat) and apply fine-grained materials (e.g., knitted fabric to a cushion), even from multiple references, while precisely preserving the source geometry. The colored boxes indicate the source of the appearance.}
  \label{fig:1}
\end{figure}

However, directly applying existing editing techniques to DiTs is not straightforward. The dominant paradigm for appearance transfer has been developed around U-Net architectures \cite{ronneberger2015u, rombach2022high}, which possess a natural spatial hierarchy. This has led to a split in the field: \textbf{semantic-aware methods} \cite{alaluf2024cross, go2024eye} excel at transferring patterns between similar objects by manipulating self-attention, but fail when semantic correspondence is absent. Conversely, \textbf{material transfer methods} \cite{cheng2024zest, garifullin2025materialfusion}, often relying on geometric controllers like ControlNet \cite{zhang2023adding}, use strong geometric priors to apply textures to dissimilar objects, but can struggle to integrate the new material naturally with the scene's lighting. This fragmentation highlights a reliance on U-Net's specific inductive biases.

The shift to DiTs, with their global, patch-based self-attention, presents a more fundamental challenge... A key component for any real image editing is a reliable inversion process. While classic DDIM inversion \cite{mokady2023null} suffers from inaccuracies, recent high-fidelity techniques based on \textbf{Rectified Flow (RF)} \cite{liu2022rectified, deng2024fireflow, wang2024taming} have enabled precise reconstruction. Still, their application has been limited to text-guided editing, leaving the more complex challenge of reference-based appearance transfer on DiTs unaddressed.

\begin{figure*}[ht]
  \centering
  \includegraphics[width=\linewidth]{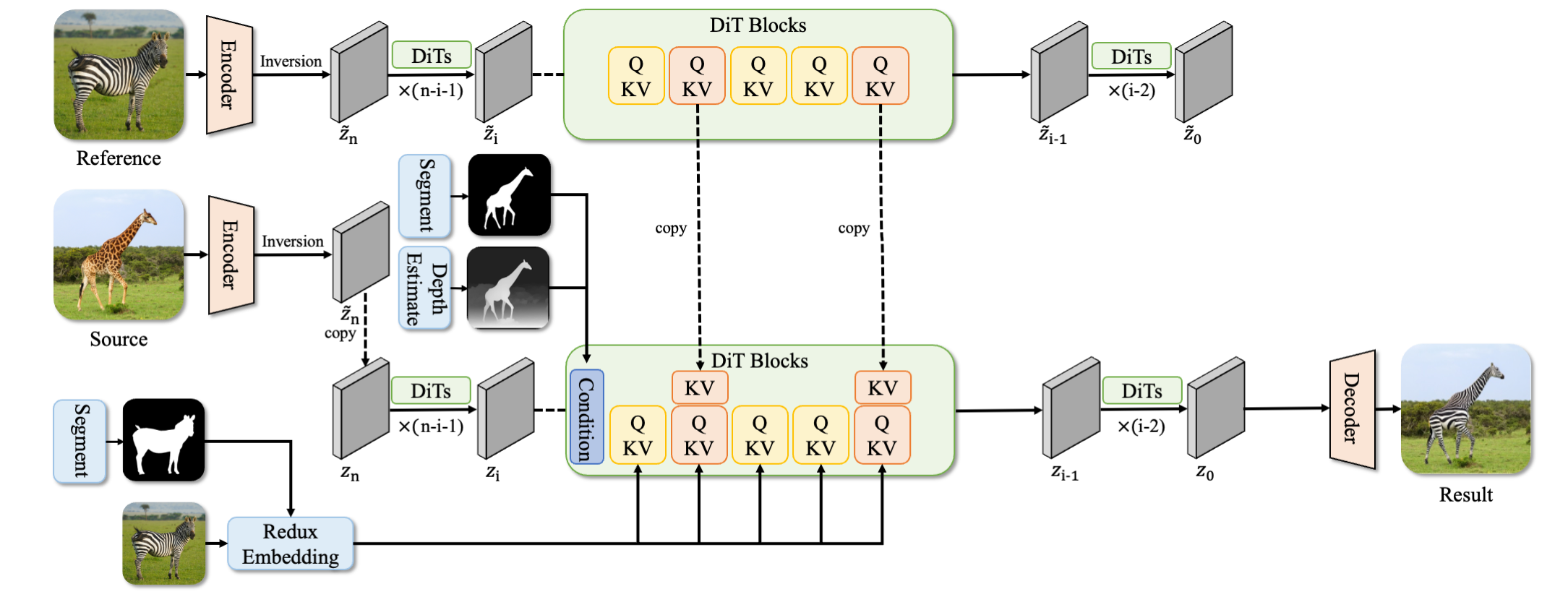}
  \caption{Overview of our framework. We perform parallel inversions on both Reference and Source images to extract appearance features (KV) and content priors ($\tilde{z}_n$), respectively. During the \textbf{Denoising Synthesis} (bottom), we initialize with the source noise $\tilde{z}_n$. For selected DiT layers (highlighted in orange), we inject appearance by expanding the attention context: copying the Reference's KV features into the Source's attention blocks. Simultaneously, a global Redux embedding and geometric conditions (Depth/Mask) guide the generation across all steps. This selective fusion strategy ensures the transfer of fine-grained appearance while strictly preserving the source structure.}
  % \caption{2}
  \label{fig:pipe}
\end{figure*}

This paper tackles this challenge head-on. We propose the first training-free, unified framework specifically designed to tame a DiT for high-fidelity appearance transfer. Our central insight is that precise geometry preservation is the key, but it is not sufficient. True fidelity requires a synergistic system that harmonizes three critical components:
\textbf{Content Prior:} We go beyond simple geometric guidance by using high-fidelity inversion to extract a detailed trajectory of the source image. This provides a strong prior for its original content, lighting, and micro-textures.
\textbf{Purified Appearance Features:} We introduce a novel disentanglement strategy to extract clean appearance features from the reference, effectively suppressing its original shape and structure.
\textbf{Guided Fusion Mechanism:} We employ a geometry-aware attention sharing mechanism that expands the attention context, allowing the model to dynamically fuse the source content prior with the reference appearance features in a controlled, localized manner.

This integrated approach allows our single, DiT-based model to handle the full spectrum of appearance transfer tasks, achieving state-of-the-art results at 1024px resolution. We demonstrate through extensive experiments that our method surpasses specialized U-Net-based approaches in both appearance fidelity and structural preservation, opening up new possibilities for controllable editing with modern foundation models.

\section{Methodology}

In this section, we detail our training-free framework for appearance transfer. Given a source image $I_{src}$ and a reference $I_{ref}$, our goal is to transfer the appearance from $I_{ref}$ to a designated region in $I_{src}$, while precisely preserving its geometry, content, and lighting. As illustrated in Figure \ref{fig:pipe}, our framework is a synergistic system of three components designed to tackle the unique challenges of editing with DiTs.

\subsection{Content Prior via High-Fidelity Inversion}

Preserving the source's structure is paramount. While geometric conditions like depth maps provide a coarse structural scaffold, they lack information about the original lighting, shadows, and micro-textures. Relying solely on them can lead to a "pasted-on" effect, where the transferred appearance feels disconnected from the scene.

To create a more holistic content prior, we leverage a high-fidelity inversion of the source image $I_{src}$. We use a Rectified Flow (RF) \cite{liu2022rectified} model, specifically FLUX.1-Depth \cite{flux2024}, and employ a higher-order solver based on FireFlow \cite{deng2024fireflow} to accurately map $I_{src}$ to its initial noise $\tilde{\mathbf{z}}_n$ through a trajectory $\{\tilde{\mathbf{z}}_0, \dots, \tilde{\mathbf{z}}_n\}$.

Instead of starting synthesis from random noise, we use a \textbf{Blended Noise Initialization} strategy. We initialize the denoising process with $\tilde{\mathbf{z}}_n$ and replay the source's trajectory up to a certain timestep $k$:
\begin{equation}
\mathbf{z}_i \leftarrow \tilde{\mathbf{z}}_i, \quad \mathrm{for }~ i = n, n-1, \dots, k.
\label{eq:replay}
\end{equation}
For timesteps $i < k$, synthesis proceeds guided by the reference appearance. This grounds the initial, high-noise generation steps in the source's authentic content, creating a content-rich canvas for the new appearance to be painted on. The hyperparameter $k$ offers a direct control over the balance between source fidelity and appearance strength.

\subsection{Purified Appearance Features via Disentanglement}

\begin{figure}[ht]
  \centering
  \includegraphics[width=\linewidth]{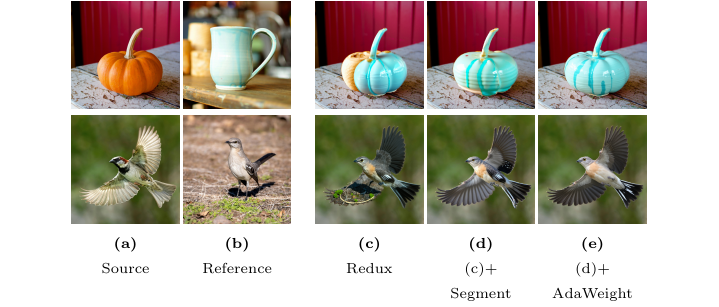}
      \caption{Comparison of different strategies for our appearance encoding module. }
  \label{fig:redux}
\end{figure}

\begin{figure}[ht!] 
  \centering
  \includegraphics[width=\linewidth]{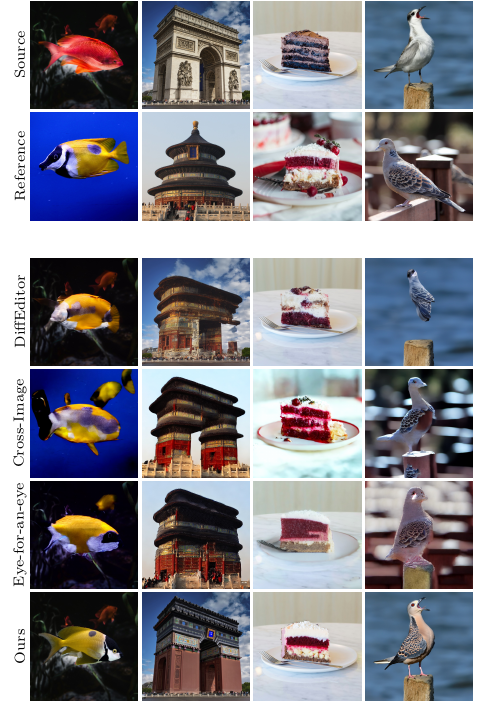} 
  \caption{Results of Semantic-Aware Transfer.}
  \label{fig:main1}
\end{figure}

To source the new appearance, we must extract features from $I_{ref}$ that are purely stylistic and free from its original structure (e.g., shape, pose). A direct encoding often leads to \textbf{structure-style entanglement}, causing the reference's shape to "leak" into the result.

To prevent this, we propose a \textbf{feature disentanglement strategy}. We use the FLUX.1 Redux adapter, but modify its encoding process. Given a foreground mask $M_{ref}$ of the subject in $I_{ref}$, we introduce a \textbf{adaptive mask-weighted embedding mechanism}. The image is processed as a sequence of patch embeddings $\{p_1, \dots, p_N\}$. Each patch $p_j$ is modulated by an adaptive weight $w_j$ proportional to its foreground mask coverage. This disrupts the encoder's ability to perceive global shape, forcing it to focus on localized, texture-like features. The result is a purified global appearance embedding $c_{ref}$ and a set of local Key/Value features, $\tilde{K}_{ref, i}$ and $\tilde{V}_{ref, i}$ (obtained via a parallel inversion of $I_{ref}$), that are rich in style but poor in structure. The results are shown in Figure \ref{fig:redux}. As illustrated, the standard Redux encoding (c) fails to decouple content, causing the reference's pose and background clutter (e.g., the ground texture) to leak into the result. In contrast, our adaptive weighting (e) effectively suppresses these structural and background cues, ensuring that only the bird's plumage pattern is transferred while the source's flight posture remains intact.

\begin{figure}[ht!] 
  \includegraphics[width=\linewidth]{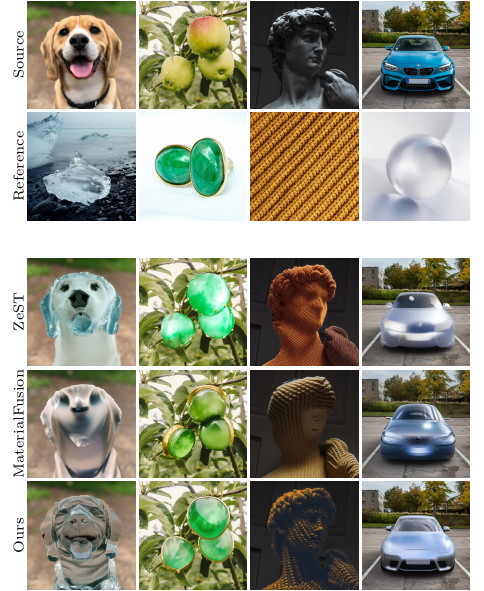}
  \caption{Results of Material Transfer.}
  \label{fig:main2}
\end{figure}

\subsection{Guided Fusion via Attention Context Expansion}

The final challenge is to seamlessly fuse the source's content prior with the reference's appearance features within a user-specified region $M_{src}$. The global embedding $c_{ref}$ sets the overall tone, but fine-grained detail transfer requires a more localized mechanism.

We introduce an \textbf{Attention Context Expansion} mechanism. Instead of replacing the DiT's attention flow, we expand it. During the denoising of the source image, for selected layers, which are first two and last two layers of both the FLUX Single-Stream and Double-Stream modules empirically\cite{avrahami2411stable}\cite{zhang2025freecus},  we leave the Query tokens $Q_{src, i}$ unchanged. However, we augment the Key and Value sequences by concatenating the features from the reference stream as the equations   
$
    K'_{src, i} = \mathrm{concat}[K_{src, i}, \tilde{K}_{ref, i}],
    \quad
    V'_{src, i} = \mathrm{concat}[V_{src, i}, \tilde{V}_{ref, i}].
$
The attention is then computed over this expanded context. This allows each query patch from the source's foreground to attend not only to its own content (via $K_{src, i}$) but also cross-image to the reference's appearance library (via $\tilde{K}_{ref, i}$). The strong constraints from our content prior ensure this fusion is seamless and confined to the target region $M_{src}$, preventing artifacts and preserving the background.

\section{Experiments}
We conduct a comprehensive set of experiments to validate our framework's effectiveness. We first detail the experimental setup, then present qualitative and quantitative comparisons against state-of-the-art specialized methods. All ablation studies are deferred to the appendix to maintain focus on our core results.

\subsection{Experimental Setup}
\noindent\textbf{Implementation Details.} Our framework is built on the FLUX.1-dev Depth model \cite{flux2024}. All results are generated at a 1024x1024 resolution. Further hyperparameter details are in the appendix.

\noindent\textbf{Compared Methods.} We compare against leading, publicly available specialized methods to ensure fair and reproducible evaluation. For semantic-aware transfer, we select DiffEditor \cite{mou2024diffeditor}, Cross-Image Attention \cite{alaluf2024cross}, and Eye-for-an-eye \cite{go2024eye}. For material transfer, we compare with ZeST \cite{cheng2024zest} and MaterialFusion \cite{garifullin2025materialfusion}. Notably, all these baselines are U-Net-based. 

\noindent\textbf{Scope of Comparison.} While recent closed-source models like Google's Gemini (formerly Nano Banana) or FLUX Kontext show impressive generative capabilities, they are not designed for the specific task of reference-based appearance transfer while preserving a source image's geometry. These models typically lack the necessary interface for multi-image, spatially-aware conditioning (e.g., source content, reference appearance, and geometric priors). A direct comparison would be inequitable, as they solve a different problem, often leading to changes in the subject's shape, pose, and background, which contradicts the core goal of our task. Therefore, we focus our comparison on methods that explicitly tackle image-guided appearance transfer.

\subsection{Qualitative Analysis}
We visually compare our unified framework against specialized methods in Figure \ref{fig:main1} (semantic-aware transfer) and Figure \ref{fig:main2} (material transfer). While presented at a reduced size for print, all results were generated at 1024px. 

\noindent\textbf{Semantic-Aware Transfer.} Figure \ref{fig:main1} presents challenging cases requiring robust structural preservation.  This highlights the effectiveness of our DiT-based approach in maintaining structural integrity during semantic transfer.

\noindent\textbf{Material Transfer.} Figure \ref{fig:main2} tests the ability to render photorealistic materials, a task that demands a deep understanding of lighting and surface properties. When applying materials with complex reflections, our method produces convincing results with realistic sheen and lighting integration. Baselines, in contrast, often yield overly smooth or plastic-like surfaces. Notably, in rendering the fine woven texture onto the David statue (Col. 3), our method succeeds where others fail, demonstrating superior fine-detail transfer on complex curvatures.

\begin{table}[ht] % Use [t] or [h] for better placement
\centering

\resizebox{\columnwidth}{!}{% % Use \columnwidth for single-column width
    \begin{tabular}{@{}l|cc|cc|c}
    \toprule
    \hline
    Model & \multicolumn{2}{c|}{Semantic-Aware Transfer} & \multicolumn{2}{c|}{Material Transfer} & {Overall} \\
    \hline
    & CLIP-I $\uparrow$ & DINO $\uparrow$ & CLIP-T $\uparrow$ & VQA $\uparrow$  & DeQA $\uparrow$ \\ 
    \hline
    DiffEditor \cite{mou2024diffeditor} & 0.7910 & 0.5269 & 0.2211 & 0.6718 & 3.1191 \\
    Cross-Image \cite{alaluf2024cross} & 0.8478 & 0.7499 & 0.2463 & 0.6084 & 3.0488  \\
    Eye-for-an-eye \cite{go2024eye} & \textbf{0.8832} & \underline{0.7521} & 0.2428 & 0.8194 &3.6758 \\
    ZeST \cite{cheng2024zest} & 0.8381 & 0.5571 & 0.2737 & \underline{0.8725}  & \underline{4.0230}\\
    MaterialFusion \cite{garifullin2025materialfusion} & 0.8064 & 0.4247 & \underline{0.2753} & 0.8598  & 3.8984 \\ 
    Ours & \underline{0.8749} & \textbf{0.8092} & \textbf{0.2927} & \textbf{0.8936} & \textbf{4.1728}  \\ \bottomrule
    \hline
    \end{tabular}%
}
\caption{Quantitative comparison on our curated dataset. Best results are in \textbf{bold}, second best are \underline{underlined}.}
\label{tab:quantitative_comparison}

\end{table}

\subsection{Quantitative Analysis}
The quantitative results on 100 image pairs are presented in Table \ref{tab:quantitative_comparison}.

\noindent\textbf{Evaluation Metrics.} We evaluate three aspects. (1) \textbf{Overall Quality} is measured by the no-reference DeQA-Score \cite{deqa_score}. (2) \textbf{Semantic Fidelity} is assessed using CLIP-I and DINO similarity. (3) \textbf{Material Fidelity} presents a unique challenge, as traditional image-similarity metrics are ill-suited for comparing a flat texture patch with a 3D object. To address this, we adopt a multi-modal evaluation protocol. We use a powerful MLLM to construct a descriptive prompt (e.g., "a photo of a \{source object\} made of \{reference material\}"). We then measure text-image alignment (CLIP-T) and conceptual correctness via VQA-Score \cite{lin2024evaluating}, they provide a reasonable and scalable method for evaluating the conceptual consistency of the material transfer, which is a critical aspect of the task.

\noindent\textbf{Analysis of Results.} The data in Table \ref{tab:quantitative_comparison} validates our framework's superiority. We achieve the highest overall quality score (DeQA-Score), confirming our visual results. In semantic transfer, our leading DINO score suggests better preservation of fine-grained details and structure. For material transfer, our significant lead across both CLIP-T and VQA-Score confirms that our method is more effective at rendering a specified material onto a new shape in a way that is both textually and conceptually coherent. These strong results across the board underscore the power of our unified, DiT-native approach.

\subsection{Ablation Study}

\begin{figure}[ht]
  \centering
  \includegraphics[width=\linewidth]{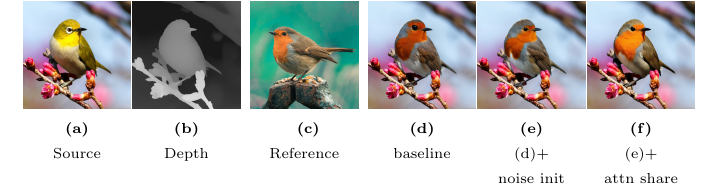}
      \caption{Step-by-step ablation of our fidelity enhancement modules. (d) Baseline (geometric guidance + Redux) roughly captures the appearance but lacks detail. (e) Adding Inversion Initialization corrects fine structural details like head/eye proportions by providing a strong content prior. (f) Our full model, with Attention Sharing, successfully transfers nuanced color and texture, demonstrating the necessity of both components.}
  \label{fig:ab}
\end{figure}

We dissect the contribution of our key fidelity enhancement modules: high-fidelity inversion and attention sharing. As shown in Figure \ref{fig:ab}: \textbf{Baseline (d):} Using only geometric guidance and the Redux encoder yields a reasonable starting point, but structural details are inaccurate (e.g., head shape) due to the lack of a rich content prior. \textbf{+ Inversion Initialization (e):} By starting the synthesis from the source's inverted noise, we significantly improve structural fidelity. The head and eye proportions now precisely match the source image. However, the appearance transfer remains incomplete, with colors and textures failing to match the reference. \textbf{+ Attention Sharing (f):} Finally, adding our attention expansion mechanism successfully injects the fine-grained color and texture details from the reference. This step-by-step improvement validates our synergistic design: high-fidelity inversion is crucial for preserving detailed structure, while attention sharing is indispensable for transferring detailed appearance.

\section{Conclusion}

In this paper, we have presented a training-free unified framework that addresses the diverse challenges within appearance transfer, from semantic attributes to physical materials. Our approach is grounded in the key insight that precise geometric preservation is the enabler for this versatility.

By combining a high-fidelity inversion process with a novel geometry-guided attention sharing mechanism, our unified framework achieves state-of-the-art, high-resolution results across both domains, showcasing its remarkable versatility without task-specific designs and opening up new possibilities for controllable, reference-based editing.

\noindent\textbf{Limitations and Future Work.}
Despite its success, our method has limitations. Our approach relies on the quality of the geometric prior. Future work could explore more robust, scale-invariant fusion strategies and extend the framework to the more challenging task of video appearance transfer.

\section{Acknowledgements}

This work was supported by the National Natural Science Foundation of China (Grant No. 62406134), Jiangsu Provincial Science \& Technology Major Project (Grant No. BG2024042), the Suzhou Key Technologies Project (Grant No. SYG2024136) and the Nanjing University-China Mobile Communications Group Co. Ltd. Joint Institute.

\bibliographystyle{IEEEbib}
\bibliography{strings,refs}

\end{document}